\documentclass[letterpaper]{article}
\makeatletter
\def\aaai@conferenceyear{2027}%
\def\copyright@on{T}
\def\showauthors@on{T}
\def\nocopyright{\gdef\copyright@on{}}
\def\copyright@year{\aaai@conferenceyear}
\def\copyright@text{Copyright \copyright\space \copyright@year,
Association for the Advancement of Artificial Intelligence (www.aaai.org).
All rights reserved.}

\gdef\copyright@on{}
\RequirePackage{iftex}
\RequirePDFTeX
\RequirePackage[T1]{fontenc}
\RequirePackage{newtxtext}
\RequirePackage{helvet}
\RequirePackage{courier}
\RequirePackage{placeins} 

 \newlength\titlebox 
 \twocolumn 
\def\addcontentsline#1#2#3{}
\def\copyrighttext#1{\gdef\copyright@on{T}\gdef\copyright@text{#1}}
\def\copyrightyear#1{\gdef\copyright@on{T}\gdef\copyright@year{#1}}

\def\maketitle{%
  \par%
  \begingroup 
    \def\thefootnote{\fnsymbol{footnote}}
    \twocolumn[\@maketitle]
    \long\def\@footnotetext##1{\insert\aaai@thanksins{%
        \protect\footnotesize\interlinepenalty\interfootnotelinepenalty
        \splittopskip\footnotesep\splitmaxdepth\dp\strutbox
        \floatingpenalty\@MM\hsize\columnwidth\@parboxrestore
        \protected@edef\@currentlabel{%
           \csname p@footnote\endcsname\@thefnmark}%
        \color@begingroup
          \@makefntext{%
            \rule\z@\footnotesep\ignorespaces##1\@finalstrut\strutbox}%
        \color@endgroup}}%
    \@thanks%
  \endgroup%
  \if T\copyright@on\insert\aaai@copyrightins{\noindent\footnotesize\copyright@text}\fi%
  \setcounter{footnote}{0}%
  \let\maketitle\relax%
  \let\@maketitle\relax%
  \gdef\@thanks{}%
  \gdef\@author{}%
  \gdef\@title{}%
  \let\thanks\relax%
}%
\long\gdef\affiliations #1{ \def \aaai@affiliations{\if T\showauthors@on#1\fi}}%
\def\@maketitle{%
  \def\aaai@theauthors{\if T\showauthors@on\@author\else Anonymous submission\fi}
  \newcounter{aaai@eqfn}\setcounter{aaai@eqfn}{0}%
  \newcounter{aaai@corrfn}\setcounter{aaai@corrfn}{0}%
  \newif\ifaaai@corrmulti\aaai@corrmultifalse
  \newsavebox{\titlearea}
  \sbox{\titlearea}{
    \let\footnote\relax\let\thanks\relax%
    \setcounter{footnote}{0}%
    \def\equalcontrib{%
      \ifnum\value{aaai@eqfn}=0%
        \footnote{These authors contributed equally.}%
        \setcounter{aaai@eqfn}{\value{footnote}}%
      \else%
        \footnotemark[\value{aaai@eqfn}]%
      \fi%
    }%
    \def\corresponding{%
      \ifnum\value{aaai@corrfn}=0%
        \footnote{Corresponding author.}%
        \setcounter{aaai@corrfn}{\value{footnote}}%
      \else%
        \footnotemark[\value{aaai@corrfn}]%
        \global\aaai@corrmultitrue
      \fi%
    }%
    \vbox{%
      \hsize\textwidth%
      \linewidth\hsize%
      \vskip 0.625in minus 0.125in%
      \centering%
      {\LARGE\bf \@title \par}%
      \vskip 0.1in plus 0.5fil minus 0.05in%
      {\Large{\textbf{\aaai@theauthors\ifhmode\\\fi}}}%
      \vskip .2em plus 0.25fil%
      {\normalsize \aaai@affiliations\ifhmode\\\fi}%
      \vskip 1em plus 2fil%
    }%
  }%
  \newlength\actualheight%
  \settoheight{\actualheight}{\usebox{\titlearea}}%
  \ifdim\actualheight>\titlebox%
    \setlength{\titlebox}{\actualheight}%
  \fi%
  \vbox to \titlebox {%
    \let\footnote\thanks\relax%
    \setcounter{footnote}{0}%
    \def\equalcontrib{%
      \ifnum\value{aaai@eqfn}=0%
        \footnote{These authors contributed equally.}%
        \setcounter{aaai@eqfn}{\value{footnote}}%
      \else%
        \footnotemark[\value{aaai@eqfn}]%
      \fi%
    }%
    \def\corresponding{%
      \ifnum\value{aaai@corrfn}=0%
        \footnote{Corresponding author\ifaaai@corrmulti{s}\fi.}%
        \setcounter{aaai@corrfn}{\value{footnote}}%
      \else%
        \footnotemark[\value{aaai@corrfn}]%
      \fi%
    }
    \hsize\textwidth%
    \linewidth\hsize%
    \vskip 0.625in minus 0.125in%
    \centering%
    {\LARGE\bf \@title \par}%
    \vskip 0.1in plus 0.5fil minus 0.05in%
    {\Large{\textbf{\aaai@theauthors\ifhmode\\\fi}}}%
    \vskip .2em plus 0.25fil%
    {\normalsize \aaai@affiliations\ifhmode\\\fi}%
    \vskip 1em plus 2fil%
  }%
}%
\renewenvironment{abstract}{%
  \centerline{\bf Abstract}%
  \vspace{0.5ex}%
  \setlength{\leftmargini}{10pt}%
  \begin{quote}%
    \small%
}{%
  \par%
  \end{quote}%
  \vskip 1ex%
}%
\def\section{\@startsection {section}{1}{\z@}{-2.0ex plus -0.5ex minus -.2ex}{3pt plus 2pt minus 1pt}{\Large\bf\centering}}
\def\subsection{\@startsection{subsection}{2}{\z@}{-2.0ex plus -0.5ex minus -.2ex}{3pt plus 2pt minus 1pt}{\large\bf\raggedright}}
\def\subsubsection{\@startsection{subparagraph}{3}{\z@}{-6pt plus -2pt minus -1pt}{-1em}{\normalsize\bf}}
\renewcommand\paragraph{\@startsection{paragraph}{4}{\z@}{-6pt plus -2pt minus -1pt}{-1em}{\normalsize\bf}}%
\skip\footins 9pt plus 4pt minus 2pt
\def\footnoterule{\kern-3pt \hrule width 5pc \kern 2.6pt }
\newinsert\aaai@thanksins%
\count\aaai@thanksins=1000%
\dimen\aaai@thanksins=\maxdimen%
\skip\aaai@thanksins=\skip\footins%
\newinsert\aaai@copyrightins%
\count\aaai@copyrightins=1000%
\dimen\aaai@copyrightins=\maxdimen%
\skip\aaai@copyrightins=\skip\footins%
\newif\ifaaai@hasfootnotes
\let\aaai@orig@makecol\@makecol
\def\@makecol{%
  \ifvoid\aaai@thanksins\else
    \setbox\footins\vbox{%
      \boxmaxdepth\@maxdepth
      \ifvoid\footins\else\unvbox\footins\fi
      \unvbox\aaai@thanksins}%
  \fi
  \ifvoid\aaai@copyrightins\else
    \setbox\footins\vbox{%
      \boxmaxdepth\@maxdepth
      \ifvoid\footins\else\unvbox\footins\fi
      \vskip 2pt
      \unvbox\aaai@copyrightins}%
  \fi
  \aaai@orig@makecol
}
\labelwidth\leftmargini\advance\labelwidth-\labelsep \labelsep 5pt
\def\@listi{\leftmargin\leftmargini}
\def\@listii{\leftmargin\leftmarginii
\labelwidth\leftmarginii\advance\labelwidth-\labelsep
\topsep 2pt plus 1pt minus 0.5pt
\parsep 1pt plus 0.5pt minus 0.5pt
\itemsep \parsep}
\def\@listiii{\leftmargin\leftmarginiii
\labelwidth\leftmarginiii\advance\labelwidth-\labelsep
\topsep 1pt plus 0.5pt minus 0.5pt
\parsep \z@
\partopsep 0.5pt plus 0pt minus 0.5pt
\itemsep \topsep}
\def\@listiv{\leftmargin\leftmarginiv
\labelwidth\leftmarginiv\advance\labelwidth-\labelsep}
\def\@listv{\leftmargin\leftmarginv
\labelwidth\leftmarginv\advance\labelwidth-\labelsep}
\def\@listvi{\leftmargin\leftmarginvi
\labelwidth\leftmarginvi\advance\labelwidth-\labelsep}
\belowdisplayskip \abovedisplayskip
\def\normalsize{\@setfontsize\normalsize\@xpt{11}}   
\def\small{\@setfontsize\small\@ixpt{10}}    
\def\footnotesize{\@setfontsize\footnotesize\@ixpt{10}}  
\def\scriptsize{\@setfontsize\scriptsize\@viipt{10}}  
\def\tiny{\@setfontsize\tiny\@vipt{7}}    
\def\large{\@setfontsize\large\@xipt{12}}    
\def\Large{\@setfontsize\Large\@xiipt{14}}    
\def\LARGE{\@setfontsize\LARGE\@xivpt{16}}    
\def\huge{\@setfontsize\huge\@xviipt{20}}    
\def\Huge{\@setfontsize\Huge\@xxpt{23}}    
\RequirePackage{xcolor} 

\AtBeginDocument{%
  \AtBeginDocument{
    \let\aaai@orig@bibliography\bibliography%
    \renewcommand{\bibliography}[1]{\FloatBarrier\aaai@orig@bibliography{#1}}%
  }
  \@ifpackageloaded{natbib}%
    {%
      \let\cite\citep
      
      \setcitestyle{aysep={}}
      \setlength\bibhang{0pt}
    }{}%
  \@ifpackageloaded{hyperref}%
    {%
      \PackageError{aaai}{You must not use hyperref in AAAI papers.}{You (or one of the packages you imported) are importing the hyperref package, which is forbidden in AAAI papers. You must remove it from the paper to proceed.}
    }{}%
  \@ifpackageloaded{bbm}%
    {%
      \PackageError{aaai}{You must not use bbm package in AAAI papers because it introduces Type 3 fonts which are forbidden.}{See https://tex.stackexchange.com/questions/479160/a-replacement-to-mathbbm1-with-type-1-fonts for possible alternatives.}
    }{}%
    \@ifpackageloaded{authblk}%
    {%
      \PackageError{aaai}{Package authblk is forbidden.}{Package authblk is forbidden. You must find an alternative.}
    }{}%
  \@ifpackageloaded{balance}%
    {%
      \PackageError{aaai}{Package balance is forbidden.}{Package balance is forbidden. You must find an alternative.}
    }{}%
  \@ifpackageloaded{CJK}%
    {%
      \PackageError{aaai}{Package CJK is forbidden.}{Package CJK is forbidden. You must find an alternative.}
    }{}%
  \@ifpackageloaded{flushend}%
    {%
      \PackageError{aaai}{Package flushend is forbidden.}{Package flushend is forbidden. You must find an alternative.}
    }{}%
  \@ifpackageloaded{fullpage}%
    {%
      \PackageError{aaai}{Package fullpage is forbidden.}{Package fullpage is forbidden. You must find an alternative.}
    }{}%
  \@ifpackageloaded{geometry}%
    {%
      \PackageError{aaai}{Package geometry is forbidden.}{Package geometry is forbidden. You must find an alternative.}
    }{}%
  \@ifpackageloaded{navigator}%
    {%
      \PackageError{aaai}{Package navigator is forbidden.}{Package navigator is forbidden. You must find an alternative.}
    }{}%
  \@ifpackageloaded{savetrees}%
    {%
      \PackageError{aaai}{Package savetrees is forbidden.}{Package savetrees is forbidden. You must find an alternative.}
    }{}%
  \@ifpackageloaded{setspace}%
    {%
      \PackageError{aaai}{Package setspace is forbidden.}{Package setspace is forbidden. You must find an alternative.}
    }{}%
  \@ifpackageloaded{stfloats}%
    {%
      \PackageError{aaai}{Package stfloats is forbidden.}{Package stfloats is forbidden. You must find an alternative.}
    }{}%
  \@ifpackageloaded{tabu}%
    {%
      \PackageError{aaai}{Package tabu is forbidden.}{Package tabu is forbidden. You must find an alternative.}
    }{}%
  \@ifpackageloaded{titlesec}%
    {%
      \PackageError{aaai}{Package titlesec is forbidden.}{Package titlesec is forbidden. You must find an alternative.}
    }{}%
  \@ifpackageloaded{tocbibind}%
    {%
      \PackageError{aaai}{Package tocbibind is forbidden.}{Package tocbibind is forbidden. You must find an alternative.}
    }{}%
  \@ifpackageloaded{ulem}%
    {%
      \PackageError{aaai}{Package ulem is forbidden.}{Package ulem is forbidden. You must find an alternative.}
    }{}%
  \@ifpackageloaded{wrapfig}%
    {%
      \PackageError{aaai}{Package wrapfig is forbidden.}{Package wrapfig is forbidden. You must find an alternative.}
    }{}%
}

\makeatother
\usepackage[hyphens]{url} 
\usepackage{graphicx} 
\usepackage{natbib} 
\usepackage{caption} 
\usepackage{booktabs}
\usepackage{amsmath}
\usepackage{amssymb}
\usepackage{array}
\usepackage{multirow}
\usepackage{xcolor}
\definecolor{revcolor}{RGB}{128,0,128}
\newcommand{\rev}[1]{\textcolor{revcolor}{#1}}
\usepackage{algorithm}
\usepackage{algpseudocode}
\usepackage{tikz}
\usepackage[most]{tcolorbox}

\definecolor{roleblue}{RGB}{35,75,115}

\definecolor{auditorblue}{RGB}{26,128,187}
\definecolor{studentblue}{RGB}{140,197,227}
\definecolor{agreegreen}{RGB}{34,150,110}
\newtcolorbox{rqbox}[1][]{
  colback=gray!5!white,
  colframe=gray!50!black,
  fonttitle=\bfseries,
  title=#1,
  title after break={},
  boxrule=0.5pt,
  arc=2pt,
  left=6pt,
  right=6pt,
  top=4pt,
  bottom=4pt
}
\definecolor{studentamber}{HTML}{E1943B}
\definecolor{controlblue}{RGB}{35,75,115}
\definecolor{docblue}{RGB}{40,90,135}
\definecolor{pdfred}{RGB}{155,60,55}
\definecolor{xlsxgreen}{RGB}{45,110,70}
\definecolor{txtgray}{RGB}{90,90,90}
\definecolor{auditorteal}{HTML}{17A589}

\newcommand{\IA}{\textsc{IntelliAudit}}

\definecolor{rqblue}{RGB}{35,75,115}
\definecolor{auditorpurple}{RGB}{105,65,135}
\definecolor{usergreen}{RGB}{45,105,70}
\definecolor{mixedteal}{RGB}{30,105,110}

\title{IntelliAudit: Using Large Language Models to Evaluate Audit Controls}
\author{
\begin{tabular}[t]{c@{\hspace{1.8em}}c@{\hspace{1.8em}}c@{\hspace{1.8em}}c}
Allison Wilson\textsuperscript{\rm 1} & Sina Moradi Sabet\textsuperscript{\rm 2} & Diar Shakimov\textsuperscript{\rm 1} & Panteha Shahrivar\textsuperscript{\rm 1} \\
{\normalfont\normalsize allison\_wilson@sfu.ca} & {\normalfont\normalsize sina.moradi@aut.ac.ir} & {\normalfont\normalsize diar\_shakimov@sfu.ca} & {\normalfont\normalsize panteha\_shahrivar@sfu.ca} \\
\end{tabular}\\[0.6em]
\begin{tabular}[t]{c@{\hspace{1.8em}}c@{\hspace{1.8em}}c}
Mohammad Reza Bagheri\textsuperscript{\rm 4} & Dean Konenkamp\textsuperscript{\rm 3} & Mohammad A. Tayebi\textsuperscript{\rm 1} \\
{\normalfont\normalsize reza.bagheri@telus.com} & {\normalfont\normalsize dean@coca-cola.com} & {\normalfont\normalsize tayebi@sfu.ca} \\
\end{tabular}
}
\affiliations{
\begin{tabular}[t]{@{}c@{\hspace{1.2em}}c@{\hspace{1.2em}}c@{\hspace{1.2em}}c@{}}
\textsuperscript{1}Simon Fraser University & \textsuperscript{\rm 2}Amirkabir University of Technology & \textsuperscript{\rm 3}Coca-Cola & \textsuperscript{\rm 4}TELUS \\
Burnaby, Canada & Tehran, Iran & Atlanta, USA & Vancouver, Canada \\
\end{tabular}
}
\begin{document}

 \maketitle

\begin{abstract}

IT audits require auditors to judge whether heterogeneous organizational evidence satisfies semantic security and compliance controls. This judgment is difficult to automate because relevant evidence is distributed across policies, records, spreadsheets, and operational artifacts, and because audit conclusions depend on evidentiary sufficiency rather than keyword matching. We present \IA, a retrieval-grounded multi-agent system for IT audit evidence evaluation. Given a control and an evidence corpus, \IA\ retrieves relevant artifacts, generates an evidence-grounded assessment, challenges adverse findings, adjudicates disagreements, and produces an auditor-facing recommendation with cited evidence, rationale, missing-evidence analysis, and remediation guidance. We instantiate \IA\ on ISO/IEC 27001 and evaluate it across multiple simulated organizations using expert auditor review and audit-readiness user feedback. The evaluation shows that \IA\ can support control interpretation, evidence-grounded reasoning, and audit-preparation workflows, while also revealing the importance of human oversight for calibrating sufficiency judgments and correcting overly permissive recommendations. These results suggest that retrieval-grounded multi-agent systems can assist audit evidence review, but should remain decision-support tools rather than autonomous certification systems.
\end{abstract}
\section{Introduction}

IT audits verify whether an organization's controls adequately mitigate its security and compliance risks. Organizations design these controls themselves or tailor them from frameworks such as ISO/IEC 27001 \cite{iso27001:2022}, hereafter ISO 27001, SOC 2 \cite{soc2tsc:2022}, the NIST Cybersecurity Framework \cite{nistcsf:2024}, and PCI DSS \cite{pcidss:2024}; auditors then independently assess them. Unlike checklist matching, this requires understanding the risk each control addresses, interpreting its semantic requirements, and determining whether heterogeneous organizational evidence sufficiently supports an audit conclusion.

This work remains labor-intensive and depends on scarce expertise. ISC2 estimates a global cybersecurity workforce gap of 4.8 million \cite{isc2:2024} and identifies governance, risk, and compliance among the hardest skill areas to fill \cite{isc2:2025}; similarly reports a U.S. shortfall of roughly 265,000 cybersecurity workers, with only enough talent to fill 83\% of employer demand. \cite{cyberseek:2025}. AI-assisted evidence review could help auditors manage this growing workload.

However, audit evidence evaluation is a high-accountability task. It determines whether a control is suitably designed and whether it operates effectively, so unsupported claims or premature compliance judgments can mislead auditors and organizations. A useful system must ground conclusions in retrieved evidence, recognize partial or missing support, expose its reasoning, and allow human auditors to challenge its recommendations.

LLM-based auditing research has concentrated primarily on financial audits \cite{wang2026auditflow, wang2025financialaudit, wang2026finauditing}, while emerging work in IT auditing targets narrower but important tasks, including audit planning, password--policy checking, and policy-to-control assessment. To the best of our knowledge, however, no prior study has introduced an AI-driven workflow for assessing whether heterogeneous evidence bundles sufficiently satisfy semantic IT-audit controls. 

To address this gap, we present \IA, a retrieval-grounded, selectively escalated multi-agent framework for IT audit evidence evaluation. The Auditor produces the initial evidence-grounded assessment, the Defender searches for overlooked mitigating evidence in adverse cases, and the Judge resolves disagreements and incorporates human feedback. Together, these agents produce traceable recommendations, identify evidentiary gaps, and suggest follow-up actions without making certification decisions.

We investigate four research questions: \textbf{(RQ1) control understanding} beyond surface text; \textbf{(RQ2) evidence sufficiency} for audit conclusions; \textbf{(RQ3) cross-artifact reasoning} across processed text, PDF, and spreadsheet evidence; and \textbf{(RQ4) audit-readiness usefulness} for users preparing for an audit.

Because no public benchmark exists for semantic IT-audit evidence evaluation, we worked with audit experts to build evidence for four simulated organizations covering 14 ISO 27001 controls. We evaluated \IA\ with practicing auditors, who assessed control interpretation and evidence use, and cybersecurity-informed audit-readiness users, who assessed audit-preparation usefulness.

The results show that \IA\ supports control interpretation and artifact-level evidence review, while sufficiency judgments over evidence bundles remain harder. Participants found its explanations useful for identifying gaps and next steps, and auditor feedback corrected overly permissive recommendations. Although evaluated on ISO 27001, \IA\ can generalize to other standards by encoding their controls and evidence requirements.

Our contributions are threefold:
\begin{itemize}
    \item We formulate IT audit evidence evaluation as an explainable, evidence-grounded recommendation task rather than autonomous certification.
    \item We introduce a selectively escalated multi-agent workflow combining retrieval, primary assessment, adversarial review, adjudication, and human feedback.
    \item We construct an audit-realistic ISO/IEC 27001 benchmark and evaluate the system through Group A: practicing auditors and Group B: cybersecurity-informed audit-readiness users assessments.
\end{itemize}

\section{Related Work}
\label{sec:related-work}

\paragraph{LLMs for compliance and audit.}
LLMs support compliance question answering, audit planning, and regulatory assessment. Prior systems combine multi-agent retrieval with regulatory knowledge \cite{agarwal2025ragulating}, support industrial supply-chain audits \cite{yao2024smartaudit}, and automate ISO 27001 auditing \cite{brachten2024iso27001}. Other work examines LLM readiness for cybersecurity governance \cite{mcintosh2024cobit}, password-policy auditing \cite{chin2025}, policy-context graph alignment \cite{chung2025}, and GDPR agreement checking \cite{amaral2023nlp}. Unlike these approaches, \IA\ evaluates the sufficiency of heterogeneous evidence for semantic ISO 27001 controls.

\paragraph{Retrieval-augmented compliance systems.}
RAG grounds outputs in external evidence \cite{lewis2021, gao2024}, with extensions for agentic reasoning \cite{singh2026} and multimodal inputs \cite{abootorabi2025}. Regulated-domain applications include risk-and-quality queries \cite{hillebrand2024advancing} and FDA-guidance assessment \cite{waikar2026rag}. \IA\ extends RAG to adversarial multi-agent control assessment with evidence-linked explanations.

\paragraph{LLMs for financial audit.}
Financial-audit benchmarks expose limitations in taxonomy-aware reasoning, cross-document consistency, standards citation, and end-to-end execution \cite{wang2026finauditing, wang2025financialaudit}. AuditFlow combines agents grounded in US-GAAP and XBRL graphs with deterministic numerical checks \cite{wang2026auditflow}. \IA\ instead addresses semantic IT controls that resist deterministic verification and adds adversarial review, adjudication, and auditor feedback.

\paragraph{Human-in-the-loop LLM auditing.}
Human--AI collaboration can uncover failures missed by either alone \cite{rastogi2023adatest}. LLMAuditor combines LLM-generated probes with human verification \citep{amirizaniani2024llmauditor}, while broader work studies feedback integration in human-agent systems \cite{zou2026}. In \IA, auditor feedback is routed through the Judge to revise traceable recommendations rather than overwrite them.

\paragraph{Multi-agent reasoning and document understanding.}
Multi-agent debate elicits and adjudicates competing interpretations \cite{du2023multiagent}, while surveys emphasize role specialization and structured coordination \cite{tran2025, guo2024}. D3 implements advocate, judge, and jury roles \citep{harrasse2026d3}; \IA\ adapts this pattern through a Defender that challenges adverse findings and a Judge that resolves disagreements. Its benchmark also connects to layout-aware document understanding, including LayoutLM-style models \cite{huang2022, xu2022, xu2020} and DocLLM \cite{wang2024docllm}.

\section{Methodology}
\label{sec:methodology}

We formulate IT audit evidence evaluation as an explainable
recommendation task and instantiate it through a
retrieval-grounded, selectively escalated multi-agent workflow.
The formulation specifies what must be produced for each control,
while the workflow describes how \IA\ retrieves evidence,
evaluates its sufficiency, reviews adverse findings, resolves
disagreements, and incorporates human feedback.

\begin{figure}[t]
    \centering
    \includegraphics[width=\linewidth]{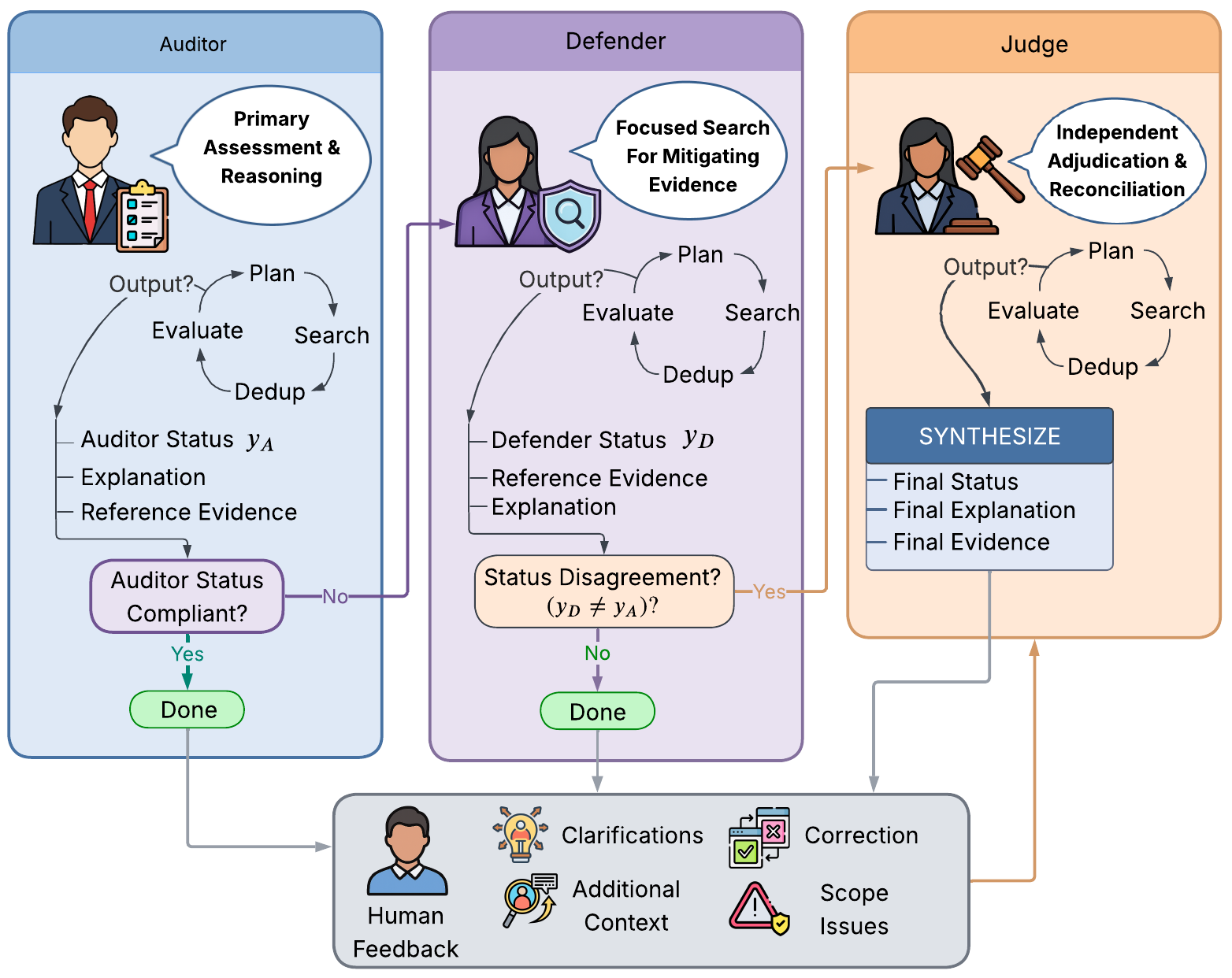}
    \caption{\IA\ workflow for assessment, counterevidence review,
adjudication, and human feedback.}
    \label{fig:intelliaudit-method-overview}
\end{figure}

\subsection{Task Formulation}

Let $R=\{r_1,\ldots,r_n\}$ denote the requirements in an audit
framework, and let $\mathcal{E}$ denote the
engagement-specific evidence corpus. Requirements may depend on
one another. We represent prerequisite relationships as a
directed graph $G=(R,\mathcal{P})$, where
$(r_i,r_j)\in\mathcal{P}$ means that requirement $r_i$ must be
satisfied before $r_j$ can be meaningfully evaluated.
Equivalently, $G$ can be encoded by a dependency matrix
$D\in\{0,1\}^{n\times n}$, where $D_{ij}=1$ iff
$(r_i,r_j)\in\mathcal{P}$.

Let $S\subseteq R$ denote the requirements whose prerequisites
have been established in the current audit state. A requirement
$r_j$ is evaluable iff
\[
\forall r_i\in R:\ 
(r_i,r_j)\in\mathcal{P}\Rightarrow r_i\in S.
\]
Non-evaluable requirements are surfaced as prerequisite gaps
rather than assigned a substantive compliance status.

Each requirement is associated with an admissible evidence
profile through the mapping
\[
\alpha:R\rightarrow 2^{\mathcal{A}},
\]
where $\mathcal{A}$ is the universe of audit artifacts, including
policies, procedures, logs, documents, PDFs, and spreadsheets.
For a target requirement $r_j$, $\alpha(r_j)$ specifies the
evidence types and criteria that the system should seek and cite.

Given an evaluable requirement $r_j$, evidence corpus
$\mathcal{E}$, and audit state $S$, \IA\ approximates the
audit-assistance function
\[
F(r_j,\mathcal{E},S)
\rightarrow
(\hat{y}_j,\mathcal{B}_j,\pi_j),
\]
where
\[
\begin{gathered}
\hat{y}_j\in
\{\texttt{COMPLIANT},\ \texttt{PARTIAL},\\
\texttt{NON\_COMPLIANT},\
\texttt{INSUFFICIENT\_EVIDENCE}\}.
\end{gathered}
\]
Here, $\hat{y}_j$ is a provisional auditor-facing
recommendation, $\mathcal{B}_j\subseteq\mathcal{E}$ is the cited
evidence bundle, and $\pi_j$ is an explanation that maps evidence
to criteria, identifies gaps or conflicts, and recommends
follow-up actions.

\texttt{PARTIAL} indicates that the evidence supports some
criteria while leaving others unmet or unsupported.
\texttt{NON\_COMPLIANT} indicates that the cited record supports
a substantive conclusion that the control is not satisfied.
By contrast, \texttt{INSUFFICIENT\_EVIDENCE} is an epistemic
abstention indicating that the available evidence is too
incomplete, conflicting, or inconclusive to support a reliable
substantive determination. In the current prototype, this
abstention is restricted to cases that reach the Judge through
agent disagreement or human feedback. The goal is therefore not
autonomous certification, but an inspectable recommendation
package for human auditor review.

\subsection{Retrieval-Grounded Control Assessment}

For each target requirement, \IA\ translates the evidence profile
$\alpha(r_j)$ into a requirement-specific search space:
\[
\begin{aligned}
\mathcal{E}_j=\{e\in\mathcal{E}: {}&
e \text{ may support at least one}\\
&\text{criterion in }\alpha(r_j)\}.
\end{aligned}
\]
The system then performs iterative retrieval over
$\mathcal{E}_j$. Each iteration identifies an evidence need,
issues a targeted query, retrieves candidate artifacts, removes
duplicates, and evaluates whether the current bundle is
sufficient for assessment. Synthesis is restricted to the
retrieved and cited bundle, reducing the risk that a status is
assigned solely from the control text or unsupported model
priors.

The assessment output is required to satisfy three traceability
constraints. First, every substantive claim in $\pi_j$ must be
supported by at least one cited artifact in $\mathcal{B}_j$.
Second, missing or ambiguous evidence must be reported explicitly
rather than treated as implicit compliance. Third, the
recommended status $\hat{y}_j$ must follow from the explanation:
the cited evidence and rationale must jointly justify the label.

\subsection{Multi-Agent Review}

\IA\ implements selective escalation rather than mandatory
multi-agent debate. The Auditor always produces the initial
evidence-grounded recommendation, while additional review stages
are invoked only when an assessment requires counterevidence
search, adjudication, abstention, or incorporation of auditor
feedback. Table~\ref{tab:review-responsibilities} summarizes the
resulting division of responsibilities.

\begin{table}[t]
\centering
\small
\setlength{\tabcolsep}{3pt}
\renewcommand{\arraystretch}{1.12}
\begin{tabular}{
@{}
>{\raggedright\arraybackslash}p{0.30\columnwidth}
>{\raggedright\arraybackslash}p{0.64\columnwidth}
@{}}
\toprule
\textbf{Risk or trigger}
& \textbf{Assigned review action} \\
\midrule

Premature adverse finding
& Defender:
Searches for overlooked mitigating or compensating evidence. \\
\addlinespace[2pt]

Conflicting interpretations
& Judge:
Adjudicates the competing interpretations using their combined
evidence bundle. \\
\addlinespace[2pt]

Inconclusive evidence
& Judge:
Issues \texttt{INSUFFICIENT\_EVIDENCE} when the record cannot
support a reliable substantive status. \\
\addlinespace[2pt]

Optimistic or incorrect finding
& Human auditor:
Challenges the provisional recommendation and triggers
re-adjudication. \\
\addlinespace[2pt]

Unsupported reasoning
& All stages:
Preserve cited evidence and explanation provenance. \\

\bottomrule
\end{tabular}
\caption{Agents review responsibilities by audit risk.}
\label{tab:review-responsibilities}
\end{table}

This division is asymmetric by design. The Defender guards against premature adverse findings, while optimistic or contextually incorrect recommendations remain provisional and subject to human review. The Judge resolves conflicting interpretations and epistemic uncertainty, and every stage preserves the supporting evidence trail.

\paragraph{Auditor.}
The Auditor performs the primary assessment. Given $r_j$ and
$\mathcal{E}$, it retrieves evidence and returns
\[
(\hat{y}_A,\mathcal{B}_A,\pi_A).
\]
The Auditor is a position-taking agent and must select a
provisional substantive status:
\[
\begin{aligned}
\mathcal{Y}^{\mathrm{sub}}
  &=\{\texttt{COMPLIANT},\texttt{PARTIAL},\\
  &\phantom{=\{}\texttt{NON\_COMPLIANT}\},\\
\hat{y}_A&\in\mathcal{Y}^{\mathrm{sub}}.
\end{aligned}
\]
It identifies satisfied criteria, missing evidence, and unresolved
uncertainty in $\pi_A$. When the evidence supports some criteria
but not others, the Auditor assigns \texttt{PARTIAL}. It cannot
assign the adjudicative abstention
\texttt{INSUFFICIENT\_EVIDENCE}; uncertainty must instead be
documented in its explanation for potential escalation or human
review.

\paragraph{Defender.}
The Defender is invoked only for adverse Auditor findings:
\[
\hat{y}_A\in\mathcal{Y}^{-},\qquad
\mathcal{Y}^{-}
=\{\texttt{PARTIAL},\texttt{NON\_COMPLIANT}\}.
\]
Its role is to determine whether the Auditor overlooked
compensating, mitigating, or differently located evidence. Given
$(r_j,\mathcal{B}_A,\pi_A,\hat{y}_A)$, the Defender conducts a
focused retrieval pass over underexplored parts of the evidence
profile and returns
\[
(\hat{y}_D,\mathcal{B}_D,\pi_D).
\]
Like the Auditor, the Defender must propose a status in
$\mathcal{Y}^{\mathrm{sub}}$ and cannot assign
\texttt{INSUFFICIENT\_EVIDENCE}. The Defender therefore does not
advocate for compliance unconditionally; it tests whether an
adverse recommendation remains justified after a targeted search
for mitigating evidence.

\paragraph{Judge.}
During automated review, the Judge is invoked when the Auditor
and Defender disagree. It is also invoked when a human auditor
provides feedback. For an agent disagreement, it receives the
competing rationales and the combined evidence bundle
\[
\mathcal{B}_{AD}
=\mathrm{DEDUP}(\mathcal{B}_A\cup\mathcal{B}_D),
\]
and produces
\[
(\hat{y}_J,\mathcal{B}_J,\pi_J).
\]
The Judge is the only agent permitted to return any status in the
full label set, including
\texttt{INSUFFICIENT\_EVIDENCE}. It selects a substantive status
when the combined evidence supports one; otherwise, it may
abstain when neither interpretation can be justified reliably.

This distinction separates \texttt{PARTIAL}, a substantive
judgment that some criteria are supported while others remain
unmet, from \texttt{INSUFFICIENT\_EVIDENCE}, an epistemic
judgment that the available record cannot sustain a reliable
substantive conclusion. Reserving abstention for the Judge allows
the Defender to search for overlooked evidence before the record
is declared indeterminate. When the Auditor and Defender agree,
their evidence and explanations are merged without further
adjudication, preserving selective escalation.

\subsection{Human Feedback}
Every \IA\ recommendation remains provisional and subject to human auditor review. An auditor may provide feedback $h_j$ to flag missing evidence, challenge a status label, correct scope assumptions, or add audit-specific context.

Rather than applying $h_j$ as a manual overwrite, \IA\ treats it as an external escalation signal and routes it to the Judge. The Judge re-evaluates the previous recommendation, cited evidence, and auditor feedback, and returns:
\[
(\hat{y}_{H},\mathcal{B}_{H},\pi_{H}).
\]
The resulting tuple may preserve or revise the previous recommendation. When the feedback reveals an evidentiary gap or misinterpretation, the Judge updates the status, evidence bundle, or explanation accordingly. This preserves a traceable connection between the original assessment, the auditor's challenge, and the resulting recommendation.

\subsection{Control Assessment Procedure}
Algorithm~1 summarizes the control-level procedure. The Auditor always runs; the Defender is invoked only for adverse findings where overlooked mitigating evidence could affect the recommendation. The Judge is invoked when the agents disagree or when human feedback is supplied. If the Auditor and Defender agree, their evidence and explanations are merged without adjudication. The procedure is selective and auditable.  Additional review occurs only under defined escalation conditions, revisions retain supporting evidence, and the final audit decision remains with the human auditor.

\begin{algorithm}[t]
\caption{\IA\ Control Assessment}
\label{alg:intelliaudit-control-assessment}
\begin{algorithmic}[1]
\Require Evaluable requirement $r_j$; evidence corpus $\mathcal{E}$
\Ensure Recommendation $\hat{y}_j$, evidence bundle $\mathcal{B}_j$, explanation $\pi_j$

\State $(\hat{y}_A,\mathcal{B}_A,\pi_A) \gets \textsc{Auditor}(r_j,\mathcal{E})$
\If{$\hat{y}_A \notin \mathcal{Y}^{-}$}
    \State $(\hat{y},\mathcal{B},\pi) \gets (\hat{y}_A,\mathcal{B}_A,\pi_A)$
\Else
    \State $(\hat{y}_D,\mathcal{B}_D,\pi_D) \gets \textsc{Defender}(r_j,\mathcal{E},\mathcal{B}_A,\pi_A)$
    \State $\mathcal{B}_{AD}\gets\mathrm{DEDUP}(\mathcal{B}_A\cup\mathcal{B}_D)$
    \If{$\hat{y}_D = \hat{y}_A$}
        \State $(\hat{y},\mathcal{B},\pi)\gets(\hat{y}_A,\mathcal{B}_{AD},\mathrm{COMBINE}(\pi_A,\pi_D))$
    \Else
        \State $(\hat{y},\mathcal{B},\pi)\gets\textsc{Judge}(r_j,\mathcal{B}_{AD},\pi_A,\pi_D)$
    \EndIf
\EndIf
\If{human feedback $h_j$ is supplied}
    \State $(\hat{y},\mathcal{B},\pi)\gets\textsc{Judge}(r_j,\mathcal{B},\pi,h_j)$
\EndIf
\State \Return $(\hat{y},\mathcal{B},\pi)$
\end{algorithmic}
\end{algorithm}

Each execution returns a provisional recommendation package
$(\hat{y}_j,\mathcal{B}_j,\pi_j)$. The package is considered
traceable only when the cited evidence bundle
$\mathcal{B}_j$ supports the explanation $\pi_j$, and the
explanation, in turn, justifies the recommended status
$\hat{y}_j$. This two-step requirement is central to \IA's
design: it keeps each recommendation inspectable, contestable,
and grounded in evidence while leaving the final audit
determination to the human auditor.

\section{Implementation} 
\begin{figure}[t]
    \centering
    \includegraphics[
        width=\columnwidth,
        height=0.4\textheight,
        keepaspectratio
    ]{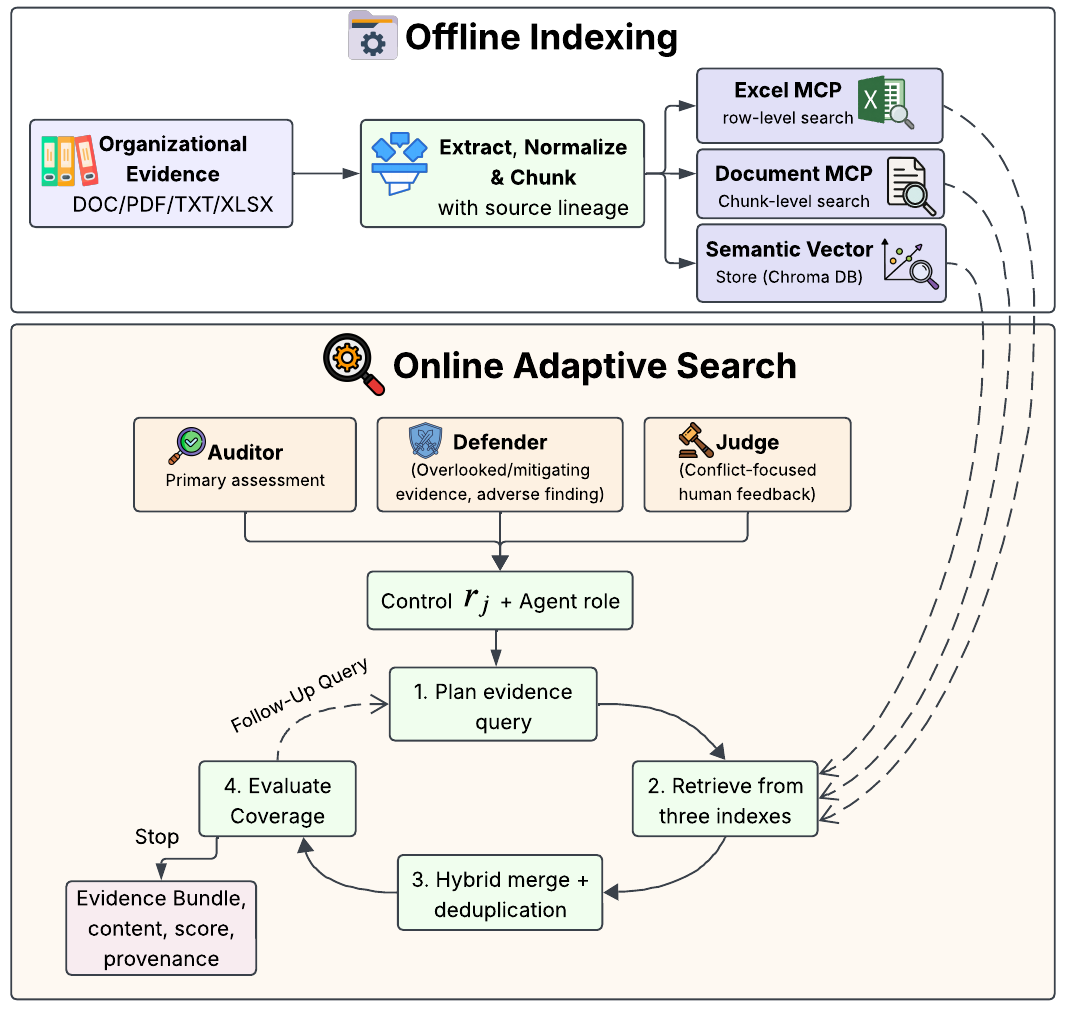}
    \caption{IntelliAudit evidence-indexing and adaptive-search architecture.}
    \label{fig:intelliaudit-indexing-search-architecture}
\end{figure}

Each control is encoded as a structured record containing its identifier, objective, evidence requirements, testing procedure, and compliance criteria. As shown in Figure~\ref{fig:intelliaudit-indexing-search-architecture}, the shared SearchAgent retrieves evidence from three complementary backends: semantic search over ChromaDB using OpenAI \texttt{text-embedding-3-small} embeddings \cite{openai:embeddings}, Document Model Context Protocol (MCP) lexical search over source-traceable document chunks, and Excel MCP lexical search over structured XLSX rows.

Documents are split into overlapping chunks with file and chunk provenance, while spreadsheets are represented as row-level records with workbook, sheet, row, and column--value metadata. Each backend returns up to $k=15$ candidates above a 0.35 threshold. Results are pooled, deduplicated by source identifier, and passed to a LangGraph plan--retrieve--deduplicate--evaluate loop.

\texttt{Claude Sonnet~4.6} performs query planning, evidence evaluation, and synthesis. Each SearchAgent call uses an initial plan plus at most three adaptive queries. The Auditor is allowed up to four iterations, and the Defender and Judge up to three each. Search stops when sufficient evidence is found, no useful follow-up query remains, or the iteration limit is reached.

\section{Benchmark and Evaluation}
We evaluate \IA\ on ISO 27001 evidence review, focusing on whether it produces evidence-supported, explainable, and audit-useful recommendations rather than automated certification decisions.
\subsection{Benchmark}
Because real audit evidence is sensitive and rarely public, we implement the benchmark across four simulated organizations. Each organization has its own evidence corpus, seeded compliance condition, and control-specific artifacts, allowing \IA\ to be evaluated across multiple audit contexts rather than a single static case. The corpus combines three sources: public standards and guidance documents, expert-crafted operational artifacts such as inventories and incident records, and synthetic policy or procedure documents generated from those seeds. Each artifact is anonymized and included in the supplementary archive with prompts and configuration files.

Each control instance is assigned one of three seeded evidence conditions—\emph{compliant}, \emph{semi-compliant}, or \emph{non-compliant}—shown in italics; system recommendation labels are shown in typewriter font. Compliant instances contain evidence satisfying the control’s semantic requirements, semi-compliant instances provide only partial support, and non-compliant instances provide no meaningful support. These conditions are distinct from \IA’s output labels: \emph{semi-compliant} typically maps to \texttt{PARTIAL}, whereas \texttt{INSUFFICIENT\_EVIDENCE} has no direct seeded counterpart.

The evaluation is organized around the four research questions introduced in the Introduction. Table~\ref{tab:rq_mapping} maps each question to its evaluation signal; evaluating 14 controls, spanning: policy, incident management, continuity, HR, and asset-management domains, with DOC, TXT, PDF, and XLSX evidence.  

\begin{table}[t]
\centering
\small
\setlength{\tabcolsep}{3pt}
\renewcommand{\arraystretch}{1.12}
\begin{tabular}{
@{}
>{\raggedright\arraybackslash}p{0.08\columnwidth}
>{\raggedright\arraybackslash}p{0.26\columnwidth}
>{\raggedright\arraybackslash}p{0.61\columnwidth}
@{}}
\toprule
\textbf{RQ}
& \textbf{Evaluation basis}
& \textbf{Evaluation signal} \\
\midrule

\textbf{RQ1}
& {Group A: practicing auditors}
& Semantic {\it understanding}: whether the recommendation interprets the control beyond a surface-level restatement. \\
\addlinespace[2pt]

\textbf{RQ2}
& {Group A: practicing auditors}
& Evidence evaluation: the {\it factual} accuracy and sufficiency of the cited evidence in {\it fulfilling} the assigned status. \\
\addlinespace[2pt]

\textbf{RQ3}
& RQ1, RQ2, and RQ4 ratings by artifact type
& Descriptive consistency of ratings across processed text, PDF, and spreadsheet evidence. \\
\addlinespace[2pt]

\textbf{RQ4}
& {Group B: Audit-readiness users}
& Practical {\it usefulness} for identifying and addressing audit-readiness gaps. \\

\bottomrule
\end{tabular}
\caption{Operationalization of the research questions through their evaluation basis and signals.}
\label{tab:rq_mapping}
\end{table}

\subsection{Human Evaluation}
For each control instance, \IA\ produces a structured recommendation containing a status label, cited evidence, rationale, missing-evidence analysis, and follow-up actions. Because no public ground-truth dataset exists for ISO 27001 evidence sufficiency, we use human evaluation as the primary reference. {Group B ratings measure usefulness for audit preparation only; no correctness or evidence-sufficiency conclusion is drawn from them, which remains the role of Group A.}

Group A consists of practicing auditors who evaluate whether the system's reasoning, evidence use, and conclusions are professionally acceptable. Group B consists of cybersecurity-informed audit-readiness users: non-auditors with cybersecurity or compliance background who {assess whether the output is understandable and actionable for audit preparation; Group B does not assess audit correctness}. Each review covers one simulated organization, and every organization is reviewed by at least two Group A auditors; participants answer 5-point Likert questions with free-text justification fields.

\subsection{Ablation and Metrics}
We compare the full workflow against an Auditor-agent-only baseline while holding the corpus, retrieval configuration, prompts, and model backbone fixed. The full pipeline invokes the Defender agent only for adverse findings and the Judge only when agents disagree or when human feedback is provided. For Judge-activation cases, reviewers complete a blinded forced-choice comparison between Auditor-agent-only and Judge-final outputs. Because these cases are selected precisely when adjudication is triggered, this ablation measures preference on contested cases rather than universal superiority of multi-agent reasoning. 

\begin{figure}[t]
    \centering
    \includegraphics[width=\columnwidth]{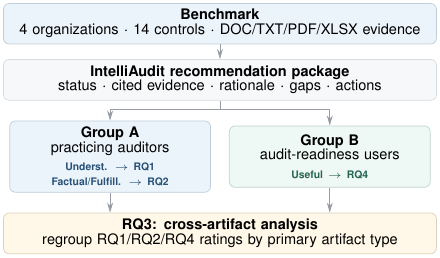}
    \caption{\IA\ evaluation design.}
    \label{fig:evaluation-overview}
\end{figure}

We report mean Likert scores, standard deviations (SD), and response counts per question ($n$). For RQ1 and RQ2, Group A participants assess control understanding {(\textit{Underst.})}, factuality of the evidence evaluation {(\textit{Factual.})}, and the evidence's fulfillment of the control {(\textit{Fulfill.}), reported per assessment unit in Table~\ref{tab:percontrol}}. For RQ3, we group RQ1, RQ2, and RQ4 ratings by primary artifact type to descriptively compare text, PDF, and spreadsheet cases. For RQ4, Group B rates {the output's usefulness (\textit{Useful.})} for identifying gaps, understanding missing evidence, and supporting remediation. Figure~\ref{fig:evaluation-overview} summarizes the evaluation design.

\section{Results}
We organize results by the four research questions. Overall, Group A found \IA's recommendations understandable and evidence-grounded, while Group B found them useful for identifying gaps and next steps. The results also show limits: sufficiency judgments were harder than control interpretation, spreadsheet-heavy cases exposed completeness errors, and human feedback often corrected overly permissive recommendations. Thus, \IA\ is best viewed as an audit-assistance system for evidence review and preparation, not autonomous certification. 

\paragraph{Control Understanding (RQ1).} Group A rated whether each recommendation reflected the semantic intent of the control rather than a surface reading of the control text. As shown in Table~\ref{tab:percontrol}, the ratings were consistently positive, with {an overall} mean of 3.92 (SD = 0.83) and 86\% of the responses equal to 4 or 5. This suggests that Group A generally found that the explanations of \IA\ were aligned with the control objectives evaluated.

The main interpretation errors were not random misunderstandings, but overly narrow assumptions about required evidence. For Control 5.1, \IA\ treated the absence of training-completion records as an acknowledgment gap, although other mechanisms may satisfy the control. For Control 5.2, it treated a responsibility matrix as effectively required, although the control requires responsibilities to be defined and allocated, not documented in a specific format. These cases suggest that \IA\ usually identifies what a control asks for, but may convert flexible audit criteria into overly specific evidence expectations. RQ1 therefore supports \IA\ for control interpretation, while motivating stronger calibration between admissible evidence examples and mandatory evidence requirements. RQ2 evaluates whether the cited evidence justifies the assigned status.

\begin{table}[t]
\centering
\footnotesize
\setlength{\tabcolsep}{0.7mm}
\begin{tabular}{@{}llcccc@{}}
\toprule
 & & RQ1 & \multicolumn{2}{c}{RQ2} & RQ4 \\
\cmidrule(lr){4-5}
Control & Artifact & Underst. & Factual & Fulfill. & Useful. \\
\midrule
5.1        & DOC  & 3.88 (0.70) & 3.38 (1.31) & 3.38 (1.02) & 3.80 (0.89) \\
5.2        & PDF  & 3.87 (0.64) & 3.75 (1.00) & 2.94 (1.39) & 4.18 (0.90) \\
5.9        & XLSX & 3.88 (0.81) & 3.53 (1.12) & 3.27 (1.16) & 3.64 (1.15) \\
5.11       & XLSX & 3.73 (1.16) & 3.94 (0.93) & 2.79 (1.67) & 4.04 (1.06) \\
5.24       & DOC  & 4.00 (0.97) & 3.76 (1.15) & 3.13 (1.64) & 4.04 (0.73) \\
5.25--27 & DOC  & 3.87 (0.99) & 3.87 (1.13) & 3.33 (1.50) & 4.17 (0.76) \\
5.29       & DOC  & 3.93 (0.73) & 3.64 (1.15) & 3.00 (1.24) & 4.08 (0.93) \\
5.30       & DOC  & 3.71 (1.14) & 3.83 (1.03) & 3.08 (1.38) & 4.08 (0.83) \\
6.4        & TXT  & 4.08 (0.51) & 3.67 (0.89) & 3.08 (1.38) & 3.83 (1.01) \\
7.14       & PDF  & 4.08 (0.51) & 4.08 (0.51) & 3.67 (0.98) & 4.00 (0.88) \\
8.1        & XLSX & 4.08 (0.76) & 4.00 (1.00) & 3.38 (1.26) & 3.08 (1.59) \\
8.4        & XLSX & 4.08 (0.79) & 3.67 (1.07) & 3.42 (1.24) & 3.08 (1.56) \\
\midrule
Overall     &      & 3.92 (0.83) & 3.75 (1.04) & 3.20 (1.32) & 3.84 (1.10) \\
\bottomrule
\end{tabular}
\caption{Per-control mean (SD) Likert ratings. RQ1--RQ2 from Group A ($n=12$--$17$ per control), RQ4 from Group B ($n=24$--$30$).}
\label{tab:percontrol}
\end{table}

\paragraph{Evidence-Grounded Assessment (RQ2).}
Group A separately rated whether \IA's evidence evaluation was factually accurate and whether the submitted evidence justified the assigned status. As shown in Table~\ref{tab:percontrol}, factual evaluation was rated higher than evidence fulfillment, with \rev{overall} means of 3.75 (SD = 1.04) and 3.20 (SD = 1.32), respectively. This distinction matters because low fulfillment scores often reflect intentionally incomplete benchmark evidence rather than system error. For Control 5.11, Group A gave a low fulfillment score because the evidence did not fully support the control, but a higher factual-evaluation score because \IA\ correctly identified evidence weaknesses. This suggests that \IA\ can often evaluate available evidence accurately even when the audit conclusion is adverse or {only partially supported}.

The main RQ2 failures involved evidence completeness and status calibration. Control 5.9 exposed a population-level spreadsheet reasoning error: \IA\ generalized from part of the asset inventory and missed other asset categories. Control 5.29 exposed a retrieval-scoping issue, where relevant incident evidence was associated with the consolidated 5.25--5.27 evidence path and was not surfaced for the disruption-control assessment. Controls 5.24 and 6.4 exposed calibration errors: \IA\ identified substantive gaps but still assigned overly permissive labels in some cases.

Overall, RQ2 shows that \IA\ is more reliable at factual evidence interpretation than at the harder audit task of converting evidence gaps into calibrated sufficiency judgments. This motivates stronger completeness checks, cross-control evidence linking, and stricter label calibration before such systems are used beyond assisted audit review.

\begin{figure}[t]
  \centering
  \begin{tikzpicture}[x=0.96cm,y=0.85cm]
    \draw[gray!55] (0,0) -- (7.40,0);
    \draw[gray!30] (0,0.000) -- (7.40,0.000);
    \node[font=\small,anchor=east] at (-0.12,0.000) {1};
    \draw[gray!30] (0,1.050) -- (7.40,1.050);
    \node[font=\small,anchor=east] at (-0.12,1.050) {2};
    \draw[gray!30] (0,2.100) -- (7.40,2.100);
    \node[font=\small,anchor=east] at (-0.12,2.100) {3};
    \draw[gray!30] (0,3.150) -- (7.40,3.150);
    \node[font=\small,anchor=east] at (-0.12,3.150) {4};
    \draw[gray!30] (0,4.200) -- (7.40,4.200);
    \node[font=\small,anchor=east] at (-0.12,4.200) {5};
    \node[font=\small,rotate=90,anchor=south] at (-0.72,2.100) {Mean rating (1--5)};
    \draw[dashed,gray!70] (0,3.150) -- (7.40,3.150);
    \fill[auditorblue] (0.925,0) rectangle (1.325,2.939);
    \draw[black,line width=0.5pt] (1.125,1.898) -- (1.125,3.980);
    \draw[black,line width=0.5pt] (1.035,3.980) -- (1.215,3.980);
    \draw[black,line width=0.5pt] (1.035,1.898) -- (1.215,1.898);
    \fill[studentamber] (1.375,0) rectangle (1.775,3.143);
    \draw[black,line width=0.5pt] (1.575,2.240) -- (1.575,4.046);
    \draw[black,line width=0.5pt] (1.485,4.046) -- (1.665,4.046);
    \draw[black,line width=0.5pt] (1.485,2.240) -- (1.665,2.240);
    \node[font=\small,anchor=north,align=center] at (1.350,-0.14) {Unstructured text\\(DOC/TXT)};
    \node[font=\small,anchor=north] at (1.350,-0.90) {$n{=}174/151$};
    \fill[auditorblue] (3.275,0) rectangle (3.675,3.074);
    \draw[black,line width=0.5pt] (3.475,2.321) -- (3.475,3.826);
    \draw[black,line width=0.5pt] (3.385,3.826) -- (3.565,3.826);
    \draw[black,line width=0.5pt] (3.385,2.321) -- (3.565,2.321);
    \fill[studentamber] (3.725,0) rectangle (4.125,3.251);
    \draw[black,line width=0.5pt] (3.925,2.315) -- (3.925,4.187);
    \draw[black,line width=0.5pt] (3.835,4.187) -- (4.015,4.187);
    \draw[black,line width=0.5pt] (3.835,2.315) -- (4.015,2.315);
    \node[font=\small,anchor=north,align=center] at (3.700,-0.14) {PDF\\\vphantom{(X)}};
    \node[font=\small,anchor=north] at (3.700,-0.90) {$n{=}55/52$};
    \fill[auditorblue] (5.625,0) rectangle (6.025,2.993);
    \draw[black,line width=0.5pt] (5.825,1.984) -- (5.825,4.003);
    \draw[black,line width=0.5pt] (5.735,4.003) -- (5.915,4.003);
    \draw[black,line width=0.5pt] (5.735,1.984) -- (5.915,1.984);
    \fill[studentamber] (6.075,0) rectangle (6.475,2.593);
    \draw[black,line width=0.5pt] (6.275,1.129) -- (6.275,4.056);
    \draw[black,line width=0.5pt] (6.185,4.056) -- (6.365,4.056);
    \draw[black,line width=0.5pt] (6.185,1.129) -- (6.365,1.129);
    \node[font=\small,anchor=north,align=center] at (6.050,-0.14) {Semi-structured\\(XLSX)};
    \node[font=\small,anchor=north] at (6.050,-0.90) {$n{=}114/98$};
    \fill[auditorblue] (0.75,4.75) rectangle (1.00,5.05);
    \node[font=\small,anchor=west] at (0.97,4.88) {Group A (RQ1+RQ2)};
    \fill[studentamber] (4.10,4.75) rectangle (4.35,5.05);
    \node[font=\small,anchor=west] at (4.38,4.88) {Group B (RQ4)};
  \end{tikzpicture}
\caption{Ratings by primary evidence type (RQ3), reported as
mean $\pm1$ SD.}
  \label{fig:rq3-multimodal}
\end{figure}
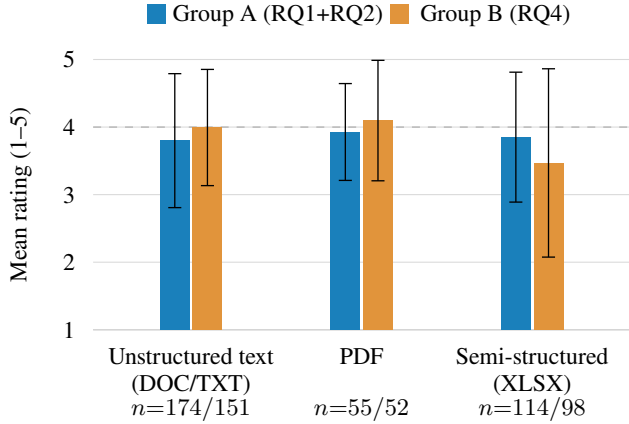

\paragraph{Reasoning Across Artifact Types (RQ3).} {RQ3 is a {\it descriptive} within-study comparison, not a modality benchmark: we group existing RQ1, RQ2, and RQ4 ratings by each case's primary evidence type, and controls and organizations were not randomized across types.} As shown in Figure~\ref{fig:rq3-multimodal}, Group A ratings were similar across processed text, PDF, and spreadsheet evidence: 3.80 (SD = 0.99, $n=174$ ratings), 3.93 (SD = 0.72, $n=55$), and 3.85 (SD = 0.96, $n=114$), respectively. Group B ratings followed the same pattern for text and PDFs, with means of 3.99 (SD = 0.86, $n=151$) and 4.10 (SD = 0.89, $n=52$), but were lower for spreadsheets at 3.47 (SD = 1.39, $n=98$).

{Within this study, \IA's outputs were rated similarly on narrative documents, processed PDFs, and structured spreadsheet records.} Spreadsheet cases reveal the main limitation: although aggregate ratings remained reasonable, Control 5.9 exposed a population-level completeness error in which \IA\ generalized from part of an asset inventory rather than checking the full spreadsheet. RQ3 provides descriptive evidence of \IA's cross-artifact applicability.

\paragraph{Audit-Preparation Usefulness (RQ4).} Group B rated whether \IA\ helped them identify deficiencies and determine what to do next. Across {the 12 assessment units (14 ISO 27001 controls; 5.25--5.27 consolidated)}, the mean usefulness rating was 3.84 (SD = 1.10, {$n = 301$ response-level ratings across four organizations)}), indicating that Group B generally found the output actionable. Ratings were strongest when the recommendation connected a concrete evidence gap to a clear remediation step, such as producing missing policy evidence, clarifying roles, or supplying additional records. {These ratings measure audit-preparation usefulness; sufficiency judgments remain with Group A.} This distinction is important because \IA\ is intended to help organizations prepare evidence and understand deficiencies, while leaving final audit conclusions to professional auditors.

\paragraph{Effect of Multi-Agent Review.} The multi-agent stages are designed for selective escalation rather than routine rewriting. Across 48 control cases (12 assessment units $\times$ 4 organizations), the Judge was activated in 5 cases (10.4\%). Among these activated cases, it changed the Auditor-only status in 3 and preserved it in 2. All changes were stricter: one moved from \texttt{PARTIAL} to \texttt{INSUFFICIENT\_EVIDENCE}, and two moved from \texttt{PARTIAL} to \texttt{NON\_COMPLIANT}; none made the recommendation more permissive. Overall, 44 of 48 final statuses (91.7\%) matched the Auditor-agent-only label; of the four non-matches, three were Judge revisions and one case produced no output.

{We further evaluated the five activation cases through a blinded forced-choice comparison (three reviewers, 15 judgments). Reviewers preferred the Judge-final output in 9 of 15 judgments, concentrated in status-changing cases (7 of 9); in the two status-preserving cases they preferred the Auditor-agent-only output (4 of 6). Given five purposively selected cases, we read this as reviewer preference on contested cases, not evidence of general multi-agent superiority.}

\paragraph{Human Feedback as Audit Calibration.}
Human feedback provides a second reliability mechanism by allowing Group A to challenge provisional recommendations. Group A interacted with the Judge in 27 case reviews (one auditor evaluating one case; six auditors, 20 distinct cases) with 15 reviews changing status after re-adjudication. Most revisions were stricter; 13 of 15 moved away from a more permissive status, and 9 of 15 began as \texttt{COMPLIANT}. No revision ended as \texttt{COMPLIANT}.

This addresses a central audit risk: unsupported positive findings. The feedback mechanism acted as a conservative calibration layer, allowing Group A to flag missing evidence, scope assumptions, or overly permissive conclusions for evidence-grounded re-adjudication rather than manual overwrite. For example, feedback on Control 5.30 revised a \texttt{COMPLIANT} finding to \texttt{INSUFFICIENT\_EVIDENCE} because continuity-testing evidence was missing; Control 6.4 similarly moved downward in all three feedback-driven re-adjudications after Group A noted the missing disciplinary policy. Thus, human oversight is part of \IA's calibration process while preserving traceability between evidence, auditor challenge, and revised recommendation.

\section{Conclusion}
\IA\ represents a step toward human-centered, AI-assisted auditing in which recommendations remain evidence-grounded, inspectable, and contestable. Our findings show the potential of selectively escalated agents to support complex audit reasoning, while reinforcing the need for professional oversight. Looking forward, audit-assistance systems should combine agentic reasoning with deterministic verification, calibrated abstention, continuous evidence linking, and structured human governance. This direction can enable scalable audit support across standards without treating AI-generated recommendations as autonomous certification decisions.

\section*{Acknowledgments}
We are grateful to the audit professionals and cybersecurity-informed participants who generously contributed their time, expertise, and feedback to our evaluation.

\clearpage
\FloatBarrier

\end{document}